\documentclass[fleqn,10pt]{wlscirep}
\usepackage[utf8]{inputenc}
\usepackage[T1]{fontenc}
\usepackage{subcaption}

\title{Fast Information Streaming Handler (FisH): A Unified Seismic Neural Network for Single Station Real-Time Earthquake Early Warning}
\author[1,2]{Tianning Zhang}
\author[1,3]{Feng Liu}
\author[1,4]{Yuming Yuan}
\author[1*]{Rui Su}
\author[1]{Wanli Ouyang}
\author[1]{Lei Bai}
\affil[1]{Shanghai Artificial Intelligence Laboratory}
\affil[2]{The Chinese University of Hong Kong}
\affil[3]{University of Science and Technology of China}
\affil[4]{Southern University of Science and Technology}
 
\affil[*]{corresponding.surui@pjlab.org.cn}

\begin{abstract}
    Earthquake early warning (EEW) systems play a crucial role in mitigating the impacts of earthquakes by providing advance warnings to areas at risk. These systems analyze seismic data to estimate the location and magnitude of earthquakes, offering valuable time for appropriate responses. However, existing EEW approaches often treat phase picking, location estimation, and magnitude estimation as separate tasks, lacking a unified framework. Additionally, most deep learning models in seismology rely on full three-component waveforms and are not suitable for real-time streaming data. To address these limitations, we propose a novel unified seismic neural network called Fast Information Streaming Handler (FisH). FisH is designed to process real-time streaming seismic data and generate simultaneous results for phase picking, location estimation, and magnitude estimation in an end-to-end fashion. By integrating these tasks within a single model, FisH simplifies the overall process and leverages the nonlinear relationships between tasks for improved performance. The FisH model utilizes RetNet as its backbone, enabling parallel processing during training and recurrent handling during inference. This capability makes FisH suitable for real-time applications, reducing latency in EEW systems. Extensive experiments conducted on the STEAD benchmark dataset provide strong validation for the effectiveness of our proposed FisH model. The results demonstrate that FisH achieves impressive performance across multiple seismic event detection and characterization tasks. Specifically, it achieves an F1 score of 0.99/0.96. Also, FisH demonstrates precise earthquake location estimation, with a location error of only 6.0km, a distance error of 2.6km, and a back-azimuth error of 19°. The model also exhibits accurate earthquake magnitude estimation, with a magnitude error of just 0.14. Additionally, FisH is capable of generating real-time estimations, providing location and magnitude estimations with a location error of 8.06km and a magnitude error of 0.18 within a mere 3 seconds after the P-wave arrives. These findings highlight the model's ability to rapidly and accurately assess seismic events The results demonstrate its potential to enhance EEW systems by providing accurate and timely information for earthquake monitoring and response.
\end{abstract}
\begin{document}

\flushbottom
\maketitle


\section*{Introduction}

     Earthquakes not only can cause significant damage to infrastructure, buildings, and natural environments, but also result in injuries and loss of life. Earthquake early warning (EEW) uses seismic monitoring instruments to detect the initial seismic waves generated by an earthquake and provide advance warning to areas that may be affected by the earthquake. When an earthquake occurs, the data collected by these monitoring systems are analyzed to determine the location, magnitude, and other characteristics of the earthquake. The purpose of an early warning system is to provide a few seconds to minutes of warning before the stronger and more destructive waves of the earthquake reach a given location. EEW is crucial for protecting lives, minimizing damage, and enhancing preparedness in the face of seismic events. While it cannot prevent earthquakes from occurring, it provides valuable time and information to respond appropriately and mitigate the impacts of these natural disasters.

     The primary steps in EEW systems involve estimating the location and magnitude of an earthquake. In some conventional EEW systems~\cite{allen2009real,wurman2007toward,serdar2014designing,chung2019optimizing,chen2015earthworm,hsiao2009development,wu2002virtual,peng2011developing}, the arrival of seismic waves at a station is detected using a method known as picker, which often employs the short-term-average/long-term-average (STA/LTA) procedure. To determine the earthquake location, a velocity model is utilized to predict the expected seismic phases. A grid search routine is then employed to estimate the earthquake location by minimizing the disparities between the observed seismic phases and those predicted by the velocity model. To estimate earthquake magnitude, some traditional EEW systems~\cite{allen2003potential,tsang2007magnitude,wurman2007toward,wu2008development,wu2008exploring,wu2006magnitude,cua2007virtual,cua2009real,bose2007earthquake,muarmureanu2011advanced,ionescu2007early,chen2015earthworm,hsiao2009development,sheen2014magnitude,sheen2017first,carranza2013earthquake,peng2011developing} analyze the amplitude (such as peak displacement) and/or the frequency characteristics (such as characteristic period) of the initial seconds of the P-wave train. This analysis involves using empirical relationships to estimate the magnitude based on the observed data. Typically, these estimates are averaged across multiple seismic stations to improve the accuracy of the magnitude estimation.

     Deep learning has emerged as a powerful tool in seismology, leveraging the abundance of seismic data accumulated over the past decades and advancements in deep learning technology. Zhu and Beroza~\cite{zhu2019phasenet} demonstrated the effectiveness of deep hybrid neural networks with multiple layers in accurately identifying seismic phases from seismic records.  Mousavi et al.~\cite{mousavi2020earthquake} introduced transformer neural networks to handle long seismic records, improving the performance of phase picking and earthquake detection. By treating single-station earthquake location and magnitude estimation as regression problems, deep learning neural networks have proven capable of estimating epicentral distance, P travel time, back azimuth, and magnitude using one-minute, three-component full waveforms~\cite{mousavi2020bayesian,mousavi2020machine}. However, current research treats phase picking, location estimation, and magnitude estimation as separate tasks, each requiring its own neural network. As a result, there is a lack of a unified neural network capable of simultaneously addressing all these EEW tasks. Moreover, most existing seismic deep learning networks rely on full three-component waveforms and are not well-suited for real-time stream data, which could significantly reduce latency in EEW systems.

     In this work, we present a novel unified seismic neural network called Fast Information Streaming Handler (FisH). FisH is designed to process real-time streaming seismic data and generate simultaneous phase picking, location estimation, and magnitude estimation results in an end-to-end fashion. By integrating these tasks within a single neural network, we simplify the overall process and enable the network to learn the nonlinear relationships between different tasks, leading to improved performance for each task. Our FisH model utilizes RetNet~\cite{sun2023retentive} as its backbone, which allows for parallel processing of long sequences during training and recurrent handling of long sequences during inference. This capability enables FisH to be applied to real-time streaming seismic data, making it suitable for earthquake early warning (EEW) systems and reducing latency. We conducted extensive experiments using the STEAD benchmark dataset~\cite{mousavi2019stanford} to evaluate the effectiveness of our FisH model.
    
\section*{Methods}

        \begin{figure}[ht]
            \centering
            \includegraphics[width=\linewidth]{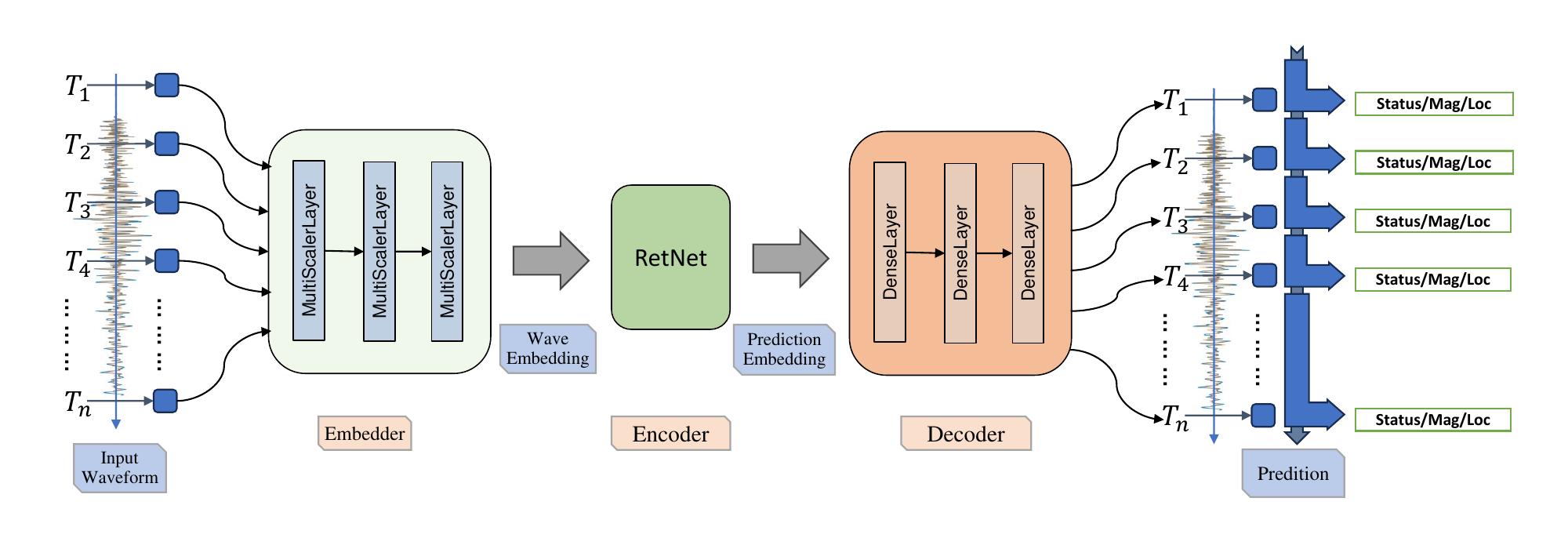}
            \caption{The overview architecture of FisH.}
            \label{fig:overview}
        \end{figure}

        As depicted in Fig.~\ref{fig:overview}, our proposed FisH model comprises three essential modules: the Embedder, Encoder, and Decoder. The Embedder module is responsible for transforming the real-time streaming waveform data at each time step into wave embeddings. These wave embeddings capture important features and patterns within the data. The Encoder module utilizes the wave embeddings to extract correlations and dependencies among them, generating prediction embeddings that encode the predictive information. Finally, the Decoder module decodes the prediction embeddings to produce the final prediction results, which include phase picking, location estimation, and magnitude estimation. By leveraging these interconnected modules, the FisH model effectively processes streaming seismic data and generates simultaneous and accurate predictions for earthquake early warning.
        
        \subsubsection*{Embedder}
            The \textbf{\textbf{Embedder}} is a critical module designed to convert various input data types into a compatible format for the \textbf{Encoder}. It takes real waveform data (Waveform) \footnote{For more advance case, it can also adapt time-varying variables (such as monitoring data outside the waveform), and non-time variables (such as site information and type)} as inputs, and transforms them into the \textbf{Embedding} which can be easily processed by the \textbf{Encoder}.
            The Embedder module is built on an intuitive design that primarily utilizes local operators to aggregate waveform information, ensuring the data is centrelized and that temporal data is effectively distilled. 
            \footnote{ If the data is not centrelized, we will apply the online \textbf{waveform decomposion} procedure, which, by using the EWM (Exponential Weighted Moving) algorithm, calculates the real-time trend and quake behavior from the original seismic waveform data.} 
            \footnote{The \textbf{\textbf{Embedder}} also incorporates symmetry in its design. This feature considers the imperceptible difference between two waveforms that differ by a sampling point, which can result in phase reversal at extremes. This is addressed by making the Convolutional kernel in the anti-symmetric, which effectively eliminates constant waveforms.} 
            
            The \textbf{Embedder} module is composed of series MultiScalerLayers (MSL) which converts the input Waveform into its WaveEmbedding. 
            $$
                \text{WaveEmbedder}=\text{MSL}_n \circ \cdots \text{MSL}_2 \circ \text{MSL}_1
            $$
            Each MSL module consists of a MultiScalerFeature (MSF), a non-linear layer (NonLi), and a normalization layer (Norm). 
            $$
                \text{MSL}_i=\text{Norm}_i \circ \text{NonLi} \circ \text{MSF}_i
            $$
            The MSF layer is a multi-perceptual aggregator composed of $N$ parallel CNN Layers and the absolute value of the entire input ABS(X), which is then compressed to a vector in dimension $D$ through a linear transformation.
            $$
                \text{MSF}_i(X) = M @[\text{CNN}_i^I(X),\text{CNN}_i^{II}(X),\text{CNN}_i^{III}(X),\text{ABS}(X)]^T
            $$
            Overall, the \textbf{Embedder} is a sophisticated module designed with advanced features to handle complex waveform data and transform it into a format that can be processed effectively, thereby enabling accurate and efficient seismic data analysis.

        \subsubsection*{Encoder}
            The Encoder module is designed to allow for our FisH to leverage the nonlinear coupled information among the wave embeddings to generate representative prediction embeddings for the Decoder module. To this end, we apply the recently proposed RetNet\cite{sun2023retentive} self-regressive architecture. The RetNet architecture is characterized by its parallel training capabilities, sequential inference with low cost, and robust performance, which are suitable for the real-time streaming seismic data and the EEW task.

            The core of the RetNet architecture is the Retention mechanism, which introduces a state matrix $s$ to map the current timestep input to output, and a specially designed matrix $a$ to propagate the state across the temporal dimension. This mechanism allows for efficient parallel training as all timestep outputs can be calculated simultaneously through matrix operations, enhancing training efficiency. In addition, the Retention mechanism can transform into recurrent neural network (RNN) sequential inference, reducing the inference cost per step to just a few matrix-vector dot operations. Another key feature of the Retention mechanism is the introduction of an exponential decay association factor $\gamma$, allowing for the control of 'memory' duration.

            In this work, we use the RetNet in real number formation and make some modifications to optimize it for the seismic phase prediction task. Generally speaking, the RetNet Backbone receive a squence embedding $[x_1, x_2, \cdots, x_n]\in R^{n\times D}$ and output the sequence prediction for each timestamp $[o_1, o_2, \cdots, o_n]\in R^{n\times D}$. Similar to Transformer, the RetNet architecture also stacks multiple identical modules $Block_i$, each of which contains a Multi-scale Retention (MSR) submodule $MSR_i$, a Feed-Forward network (FFN) submodule $FFN_i$, a Normalization module $Norm_{i}$, and a Resnet Wrapper. \footnote{By default, we perform two Normalization operations before and after the MSR module, which ensures numerical stability. }
            $$
                \text{Retnet} = \text{Block}_{N} \circ \text{Block}_{N-1} \circ \cdots \circ \text{Block}_{2}\circ \text{Block}_{1} \\
                \text{Block}_{i} = \text{FFN}_{i} \circ \text{Norm}_{i} \circ  (\text{MSR}_{i} \circ \text{Norm}_{i}+\text{Id})
            $$
            Usually, the Feed-Forward network (FFN) module is a two-layer Linear Layer sandwiched by a Layer Normalization module: $\text{FFN}_{i} = \text{FN}i^{\text{II}}\circ \text{LN} \circ\text{FN}{i}^{\text{I}}$. The non-linear transformation of Linear layer is 'gelu' and the Normalization module (Norm) is `RMSNorm`. The Multi-scale Retention (MSR) module is a multi-scale self-regressive module, consist of Rotary Position Embedding\cite{su2024roformer} and Self Retention. Same as Transformer, the Self Retention will project the input embedding into query $Q$, key $K$, value $V$ and gate $G$ state, and do the standard 'Retention' operation below following a standard Linear projection.
            $$
                O =  \text{Norm}(G)*\text{Retention}(Q, K, V)\\O = A_o O + b_o
            $$
            Here, we use the Einstein summation convention briefly review the two equivalence equations for the Retention operation. \footnote{More details can be found in supplementary material. For instance, the third equivalence formulation of retention.  }
            \begin{itemize}
                \item Parallel mode: $O_i^c=\mathcal{R}_i^{\textbf{j}} v_{\textbf{j}}^c 
                                           = \lambda^{i-\textbf{j}} \delta_{\textbf{j}\leq i} q_i^{\bf{a}}k_{\bf{a}}^{\textbf{j}} v_{\textbf{j}}^c$
                \item Recurrent mode: $O_i^c=q_i^{\bf{a}}H_{c,{\bf{a}}}^i$ and $H^{n+1}_{c,a} = \lambda H^{n}_{c,a} + k_a^{n+1}v_{n+1}^c$ and $H^{0}_{c,a}=0$
            \end{itemize}
            Notice, we omit the normalization term and stable term in the above equations. A complete math formulation can be found in the supplementary material.

        \subsubsection*{Decoder}
            Within the Decoder module of our FISH model, the task-specific downstream outputs, including phase picking results, earthquake magnitudes, and location, are generated. The Decoder consists of three downstream heads, each responsible for producing the corresponding task outputs.

            For the phase picking head, a memory bank with a temporal size of $T$ is utilized. At each time step, the prediction embedding is stored at the last position of the memory bank, and the prediction embedding at the first position is removed. This memory bank retains the prediction embeddings for the last $T$ time steps. To generate the phase picking results, several convolutional layers are employed to localize the P/S arrival positions within this memory bank. If no P/S arrival is detected within the last $T$ time steps, the output is set to 1, indicating the absence of P/S arrivals.

            For the location and magnitude estimation heads, they employ multiple linear layers to map the current prediction embedding to seismic quantities such as distances and magnitudes, enabling accurate estimation of earthquake locations and magnitudes.
        
    \subsubsection*{Training Paradigm}
        The raw data feed to the model is the waveform. We will only allow the normalization method that has online version. For example, shift or amplifier for a constant value.
        \footnote{For those non-centrelized data like DiTing, we usually apply the online waveform decomposion procedure, which, by using the EWM (Exponential Weighted Moving) algorithm, calculates the real-time trend and quake behavior from the original seismic waveform data.} Besides, we will assign the \textbf{focus range} that will the regression backward activated. For example, give a sequence $L=9000$, there is only few timestamps related to the earthquake, then we can assign a focus range from the $200$ stamps ahead from $P$-arrive to $3000$ stamps after $P$-arrive; or we can assign a full sequence focus range so the model should continues report from before earthquake to after earthquake. 
        Topical subheadings are allowed.
        
        To augment the data, we will retrieve a continue sub-sequence from the raw data. Detaily, take STEAD as an example, the raw data contain a $L=6000$ waveform sequence and the position of the $P$-arrive.
        We will first randomly calibrate the starting position of the retrieve sequence in a interval $[P-a, P]$ ahead of $P$-arrive. Then we retrieve a $L=6000$ sequence, we will padding zero or noise for those missing data. 
       
        The main training framework is called "Sea" mode. The "Sea" mode is the advance version of the Recurrent mode.
        \begin{figure}[ht]
            \centering
            \includegraphics[width=\linewidth]{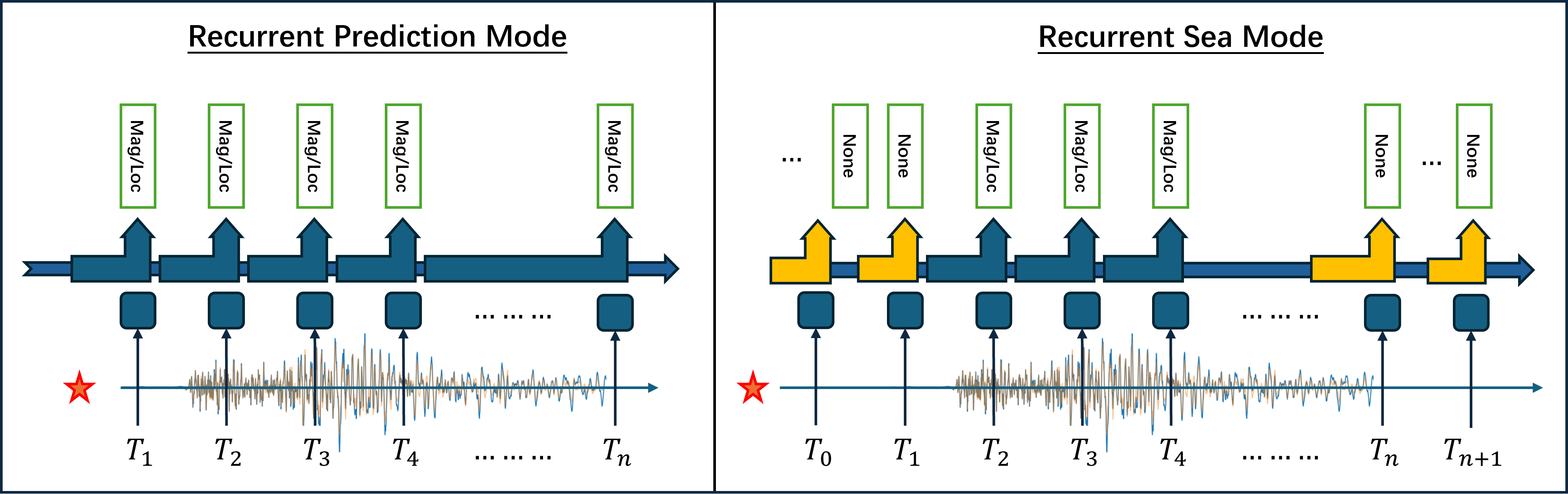}
            \label{fig:stream}
            \caption{Overview of the Recuurent Prediction Mode and the Recurrent Sea Mode.}
        \end{figure}
        In Recurrent mode, given an waveform sequence containing earthquakes, our task is to obtain whole the seismic quantity for each timestamp. The target label is also a sequence: assuming our model is $f$, the label is the magnitude of this earthquake event $\text{Mag}{true}(L,1)$ , the input sequence is $Waveform$ (L,3) and predicted value should be $\text{Mag}{pred}=F(\text{EventSeq}) (L,1)$. In this paradigm, model play the role that pack the history information into the latest embedding, and then transcribes to the seismic target.
        
        The "Sea" mode is no different from the Recurrent mode except a subtle controlling for the "hidden increment" of Retnet between the "quake" and "non-quake" states. 
        The "hidden increment" is defined as the information addition for each timestamp in Retention Network RNN mode. 
        As shown in above, we know that RNN retention handle a recurrent update formula as $O_i^c= q_i^{\bf{a}}H_{c,{\bf{a}}}^i$ and updating at each stamp via 
        $$
        H^{n+1}_{c,a} = \lambda H^{n}_{c,a} + k_a^{n+1}v_{n+1}^c
        $$
        Here $\lambda<1$ is a memory decay term, and $k_a^{n+1}v_{n+1}^c$ is the "hidden increment". Now we will require the model has the ability named "zero norm response": when the model pass a earthquake and has experienced enough long time, the state of the model should to regress back its original state. The model will remain such a inactivate status until the next earthquake. And the whole life of the model is a perfect cycle of those "quake" and "non-quake" states. Since we initialize model with $H^{0}_{c,a}=0$, we actually know the "inactivate" status is the zero state. Combine with the updating formula, we can assign 
        
        \begin{align*}
            \Delta_{n+1}=||k_a^{n+1}v_{n+1}^c|| \neq0 \quad \text{during quake}\\
            \Delta_{n+1}=||k_a^{n+1}v_{n+1}^c|| = 0 \quad \text{during noise}
        \end{align*}

        Here is serveral interesting factor after controlling "hidden increment" term 
        \begin{itemize}
            \item This model can train or inference start at any noise time, regardless of its history. 
            \item As long as the quake interval is larger than the memory setting, a model trained in single quake sequence can spontaneously work well in the multi-quake sequence.
            \item The valid inference length of the model is infinity.
        \end{itemize}

    \subsubsection*{Some Training Detail}
        We usually use `Huggingface accelerate'\cite{accelerate} framework and the `Deepspeed zero1'\cite{rasley2020deepspeed} training system to train the model under `bf16' digital precision. The training system is a 8 cards/32 cards 3090Ti/A100 system.


\section*{Results}

    To validate our FisH model, we conduct experiments and show results on the STEAD dataset and the INSTANCE dataset:
    \begin{itemize}
        \item STEAD-Full dataset\cite{mousavi2019stanford}: A Global Data Set of Seismic Signals for AI 
        \item STEAD-BDLEELSSO dataset\cite{mousavi2020bayesian}: A subset of STEAD-Full dataset. Filtering by used earthquake waveforms recorded at an epicentral distance less than 110 kilometers,signal-to-noise ratio of 25 decibels or more and data from stations where the North-South and East-West components are correctly aligned with their geographical directions.
        \item INSTANCE dataset\cite{michelini2021instance}: The Italian seismic dataset for machine learning
    \end{itemize}
    We also evaluate our proposed model in terms of four different seismic tasks: phase picking, magnitude estimation and location estimation. The experimental results on the STEAD-BDLEELSSO dataset are used as the example to analyze the performance of the proposed model.

        \subsubsection*{Phase Picking}
            \begin{figure}[ht]
                \centering
                \includegraphics[width=\textwidth]{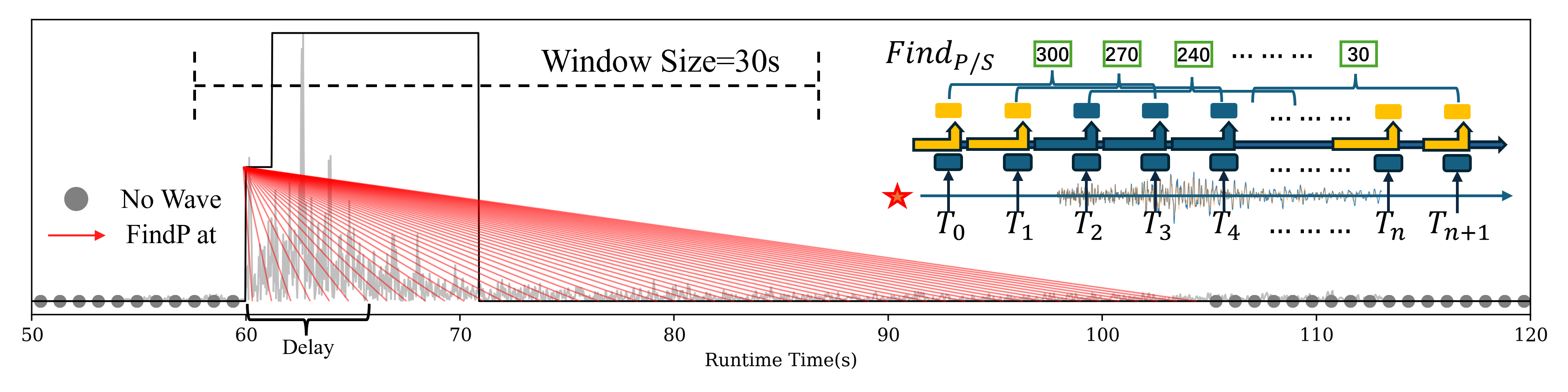}
                \caption{A Example diagram for Online P phase picking task}
                \label{FindPS_Diagram}
            \end{figure}

           To generate online phase picking results, as shown in Fig~\ref{FindPS_Diagram}, for each time step, we use the historical information of last $T$ time step in the memory bank to generate the current picking results and continuously doing so as the time goes. Generally, we set $T=30s$. Note that we use zero padding when the historical data points are missing. However, our FisH model always outputs P and S arrival positions within the memory bank, which are invalid before P/S-arrival time. We then use the testing set of the STEAD-BDLEELSSO dataset to evaluate the online real-time phase picking performance of our proposed model. Our FisH model demonstrates remarkable performance in promptly picking P/S arrival times upon the arrival of the corresponding waves. At the P/S arrival times, it achieves a precision and recall of 99\% for P arrival picking. As for S arrival picking, our model achieves an accuracy rate of 96\% and a recall rate of 90\%. These results highlight the effectiveness of our FisH model in accurately and efficiently identifying seismic wave arrivals without using the information after P/S arrivals, thereby contributing to the improvement of earthquake early warning systems.
           
           We also follow the evaluation setting in EQTransformer~\cite{mousavi2020earthquake} to evaluate the offline picking performance for our proposed model, where the complete seismic wave inputs are given and the information after P/S arrivals is used. Specifically, we use the first 30 seconds of the each seismic wave sequence from the testing set of the STEAD-BDLEELSSO dataset and generate the picking results. In this scenario, our FisH model can achieve the performance of 99\% in terms of both precision and recall on this dataset.

           By considering both online real-time phase picking and offline performance evaluations, our FISH model showcases its capability to deliver accurate and efficient phase picking results for earthquake early warning systems.
                


        \subsubsection*{Magnitude Estimation}
\begin{figure}[ht]
                \centering
                \begin{subfigure}[b]{0.8\linewidth}
                    \centering
                    \includegraphics[width=\textwidth]{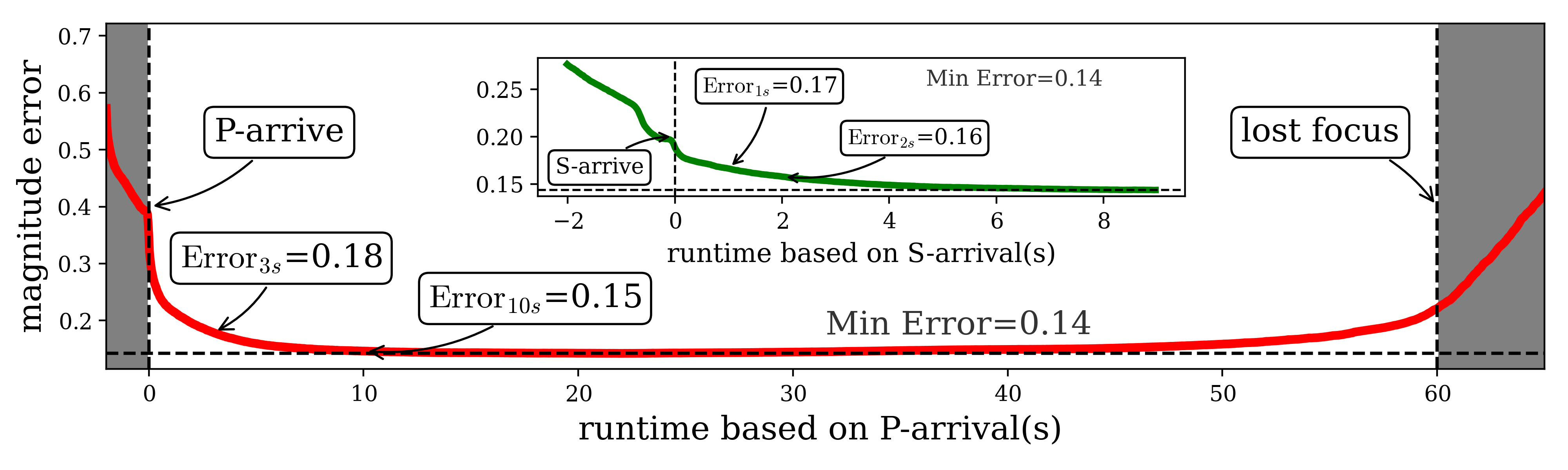}
                    \caption{The Runtime Magnitude Error for FisH}
                    \label{Magnitude_error}
                \end{subfigure}
                
                \begin{subfigure}[b]{0.8\linewidth}
                    \centering
                    \includegraphics[width=\textwidth]{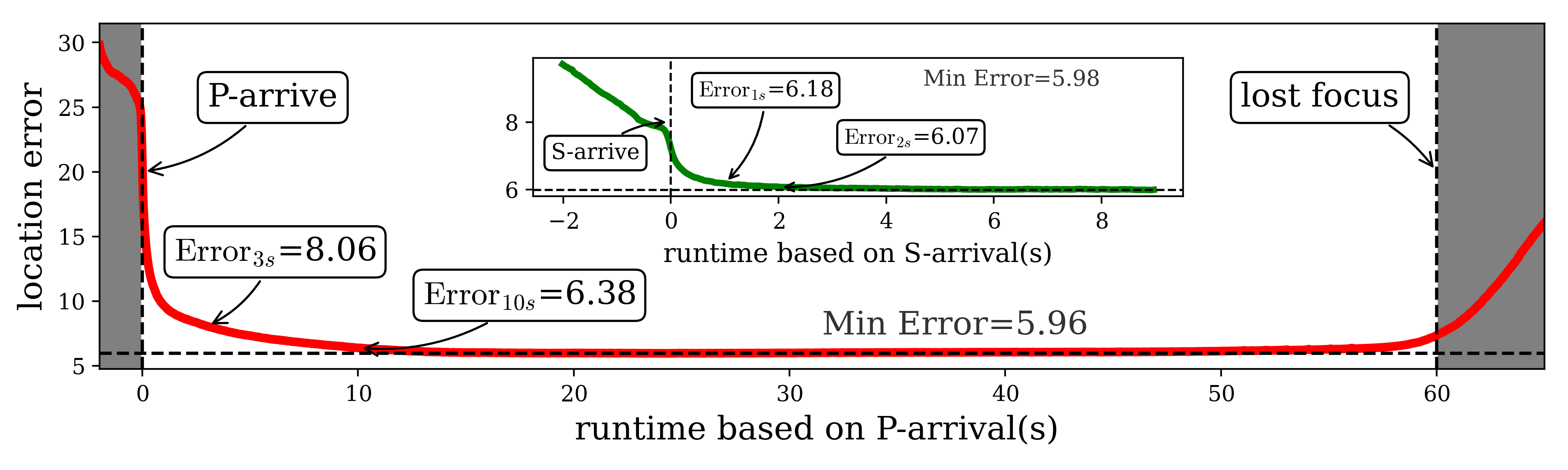}
                    \caption{The Runtime Location Error for FisH}
                    \label{Position_shift_error}
                \end{subfigure}
                \caption{The magnitude and position error of the model. The x-axis is the time after the P-arrive and the y-axis is the error.}
                \label{fig:mag_loc_error}
            \end{figure}

            In our FisH, our Decoder directly estimate earthquake magnitude from the prediction embedding at each time step. We use the P-arrival time as the reference time, and evaluate the online magnitude estimation performance by using the mean absolute error (MAE) between the estimated magnitude and the target magnitude from 2 seconds before the P-arrival time to 70 seconds after the P-arrival time. The result on the STEAD-BDLEELSSO dataset is shown in Fig.\ref{Magnitude_error}. It can be clearly seen that the error quickly drops after the P wave arrival and continues to decrease to the minimum (around 0.15) at around 7 seconds later. Notably, the FisH model achieves a very low magnitude estimation error (0.18) within 3 seconds after P wave arrives. This rapid convergence to the optimal magnitude estimate is particularly impressive and highlights the model's potential for applications in earthquake early warning systems. Inside Fig.\ref{Magnitude_error}, we also show the MAE collected from 2 seconds before the S-arrival to 9 seconds after the S-arrival. Our FisH model is able to provide a good and continuous prediction throughout the entire quake event. It can quickly achieve the best magnitude estimation 6 seconds after the S-arrival. By effectively leveraging the prediction embeddings generated by the Decoder module, our FisH model showcases its capability to estimate earthquake magnitudes with high accuracy and timeliness, further enhancing the overall performance of earthquake early warning systems.           
        \subsubsection*{Location Estimation}
            Traditionally, the location of earthquake is determined by estimating its distance and azimuth-deg to the corresponding station. In this work, our FisH model estimate the relative location of the earthquake $(x_q, y_q)$ to the corresponding to station. We use the location error to evaluate the location estimation performance of our FisH model, which is defined as the Euclidean distance between the estimated location $(x_e, y_e)$ and the target location $e=\sqrt{(x_q-x_e)^2 + (y_q - y_e)^2}$. 
            
            The location error of our estimated location from 2 seconds before the P-arrival time to 70 seconds after the P-arrival time on the STEAD-BDLEELSSO dataset is shown in Fig.\ref{Position_shift_error}. It can be seen that the location error quickly decays after the P-arrival and reaches the minimum of around 6km
            after approximately 10 seconds. Similar to the observations for the magnitude estimation, the location error of our FisH can fast converge and achieve 8.06 kilometers within 3 seconds after P-arrival. Our model has a focus limit of around 60 seconds, after which the model resets back to its initial state. Similar to the Magnitude Estimation task, we also report the location error of our estimated location using the S-arrival time as the reference time. We can clearly find that our FisH model can provide relatively accurate   location estimation even before the S wave arrival. This is a challenge task for existing single-station location estimation methods, which might need the information of s-arrival time. Additionally, when the S-arrival information is perceived the model, it gives a fast response and reaches the best performance around 1 second later. 
            \begin{figure}[ht]
                \centering
                \includegraphics[width=\textwidth]{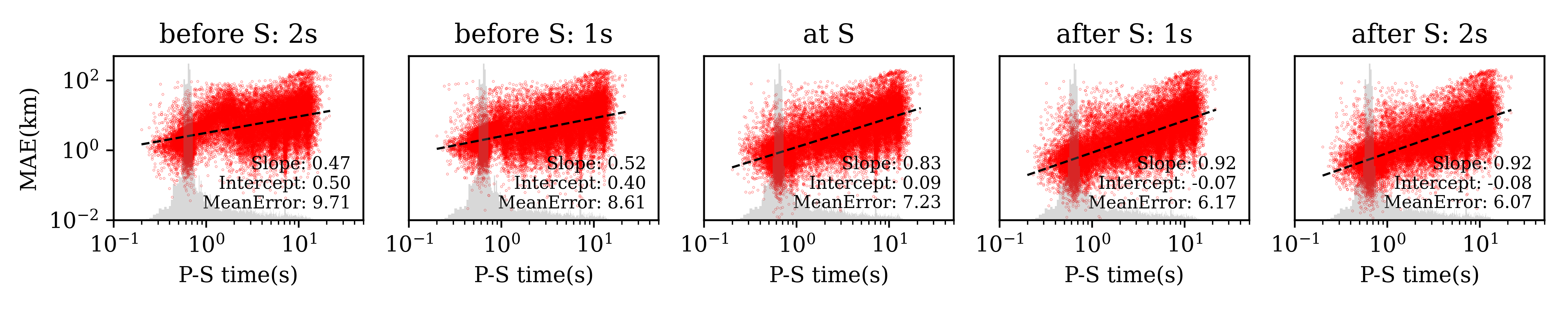}
                \caption{Shift Error Dependent to the P-S times. A clearly linear dependency in log-log plot, i.e. a power law.}\label{Error_Dependent}
            \end{figure}

            Interestingly, we find that the error of the location estimation has an obvious dependence on the interval between the P-arrival and S-arrival, which is approximately equivalent to the distance. As shown in Fig.\ref{Error_Dependent}, we plot all the test sampling points in a 2D scatter view. The x-axis is the P-S time, and the y-axis is the error of the location estimation. Note that both the x-axis and the y-axis are in log scale. The gray background represents the histogram of the P-S time. It can be clearly seen that the error of the location prediction follows a power law (linear in log-log plot) with respect to the P-S time (i.e., the source distance). This observation aligns with the intuition that the closer the source, the more accurate the prediction. However, the power law instead of a linear law is a new finding.

    \subsection*{Result in Other dataset and Compare to state-of-the-art works}
            In this section, we compare the performance of our FisH model to other state-of-the-art works in terms of the phase picking, localization estimation and magnitude estimation tasks. Additionally, to further verify the generalization ability of our FisH model, conduct experiments on another large seismic waveform benchmark dataset Instance~\cite{michelini2021instance} and report the experimental results.
            
            Most existing deep neural network-based phase picking methods primarily report their performance in an offline setting, where the complete seismic waveforms are available. To enable a fair comparison, we also evaluate the offline phase picking capabilities of our FisH model. Specifically, we use the first 30 seconds of the original seismic waveforms as input to our model and generate the corresponding P/S picks. The offline phase picking results, including precision, recall, and F1 score, are presented in Table~\ref{P-picking-performance} and Table~\ref{S-picking-performance}. Our FisH model achieves comparable or even better performance when compared to other state-of-the-art phase picking methods. These results demonstrate the effectiveness and robustness of our proposed FisH model in accurately identifying seismic phases, even in an offline setting where the full waveform information is available.
            \begin{table}[ht]
                \centering
                \begin{tabular}{|lccccc|}
                \hline
                {\textbf{Model}}                                 & {\textbf{Pr}} & {\textbf{Re}}& {\textbf{F1}}  & {\textbf{dataset}} & {\textbf{Data size}} \\ \hline
                { \textbf{FisH(Ours)}}            & { 0.99}& { 0.99}& { 0.99}   & { BDLEELSSO}         & { 0.2M} \\ 
                { \textbf{FisH(Ours)}}            & { 0.99}& { 0.98}& { 0.99}   & { STEAD-Full}        & { 1.2M}  \\
                { \textbf{FisH(Ours)}}            & { 0.98}& { 0.81}& { 0.89}   & { INSTANCE}          & { 0.8M}  \\\hline
                { \textbf{LPPN}}\cite{10.1785/0220210309}              & { 0.95}       & { 0.94}      &    { 0.94}            & { STEAD-Full}        & { 1.2M}    \\
                { \textbf{EPick}}\cite{li2022epick}              & { 0.95}       & { 0.97}      &    { 0.96}            & { STEAD-Full}        & { 1.2M}              \\
                { \textbf{CapsPhase}}\cite{9467532}              & { 0.94}       & { 0.99}      &    { 0.97}            & { STEAD-Full}        & { 1.2M}              \\
                { \textbf{EQTransformer}}\cite{zhu2019phasenet}   & { 0.99}& { 0.99}& { 0.99}   & { STEAD-Full}        & { 1.2M}  \\
                \hline
                \end{tabular}
                \caption{P-picking performance}\label{P-picking-performance}
                
            \end{table}
            \begin{table}[ht]
            \centering
            \begin{tabular}{|lccccc|}
            \hline
            {\textbf{Model}}                                 & {\textbf{Pr}} & {\textbf{Re}}& {\textbf{F1}}  & {\textbf{dataset}} & {\textbf{Data size}} \\ \hline
            { \textbf{FisH(Ours)}}                                             & { 0.96}       & { 0.95}      &    { 0.96}            & { BDLEELSSO}         & { 0.2M}              \\ 
            { \textbf{FisH(Ours)}}                                             & { 0.95}       & { 0.94}      &    {0.95}           & { STEAD-Full}        & { 1.2M}              \\
            { \textbf{FisH(Ours)}}                                             & { 0.92}       & { 0.79}      &    { 0.85}            & { INSTANCE}          & { 0.8M}              \\\hline
            { \textbf{LPPN}}\cite{10.1785/0220210309}              & { 0.83}       & { 0.84}      &    { 0.84}            & { STEAD-Full}        & { 1.2M}    \\
            { \textbf{EPick}}\cite{li2022epick}              & { 0.95}       & { 0.95}      &    { 0.95}            & { STEAD-Full}        & { 1.2M}              \\
            { \textbf{CapsPhase}}\cite{9467532}              & { 0.88}       & { 0.99}      &    { 0.93}            & { STEAD-Full}        & { 1.2M}              \\
            { \textbf{EQTransformer}}\cite{zhu2019phasenet}              & { 0.99}       & { 0.96}      &    { 0.98}            & { STEAD-Full}        & { 1.2M}              \\
            \hline                  
            \end{tabular}
            \caption{S-picking performance}\label{S-picking-performance}
            \end{table}

            Our FisH model is capable of estimating the earthquake location at every time step during the seismic event. For the final location estimation result, we select the location prediction generated at 20 seconds after the P-wave arrival time. We then report the location error, distance error, and back-azimuth angle error in Table~\ref{Location-Pred-performance}.It is important to note that our FisH model only estimates the relative location of the earthquake with respect to the observation station. It does not directly generate the distance and back-azimuth angle results. These additional parameters are calculated based on the estimated relative location provided by the FisH model.The results show that our FisH model achieves a location error of 6.0 kilometers, a distance error of 2.6 kilometers, and a back-azimuth error of 19 degrees on the Stead-BDLEELSSO dataset. This performance significantly outperforms all the compared state-of-the-art methods including SeisPairNet~\cite{meng2024seisparanet}, Mousavi~\cite{mousavi2020bayesian}, Complex CNN~\cite{ristea2021complex} and the Fourier Transformer~\cite{ge2024multi}. Furthermore, the similar results obtained by our FisH model on the STEAD-Full and Instance datasets suggest the strong generalization ability and robustness of our proposed approach.These findings demonstrate the effectiveness of the FisH model in accurately estimating the location of earthquakes using single-station seismic observations. The model's ability to provide reliable location estimates in a timely manner is crucial for enhancing the capabilities of earthquake early warning systems.
            \begin{table}[ht]
                \centering 
                \begin{tabular}{|lcccccc|}
                \hline
                {\textbf{Model}}                                & \textbf{location error} &  \textbf{distance error} &  \textbf{angle error} &  \textbf{magnitude error}  &  \textbf{dataset}  &  \textbf{Data size} \\ \hline
                { \textbf{FisH(Ours)}}                                &  6.0km             &   2.6km            &  19°             &  0.14             &   BDLEELSSO          &  0.2M               \\ 
                { \textbf{FisH(Ours)}}                                &  9.4km             &   3.9km            &  22°             &  0.15             &   STEAD-Full         &  1.2M               \\
                { \textbf{FisH(Ours)}}                                &  8.5km             &   3.6km            &  22°             &  0.15             &   INSTANCE           &  0.8M               \\\hline
                { \textbf{MagNet}}\cite{mousavi2020machine} &  -             &  -            &  -             &  0.20             &   BDLEELSSO          &  0.2M                   \\
                { \textbf{SeisPairNet}}\cite{meng2024seisparanet} &  -             &   8.85km            &  31°             &  0.29             &   BDLEELSSO          &  0.2M               \\
                { \textbf{Mousavi}}\cite{mousavi2020bayesian} &  7.3km             &   5.4km            &  33°             &  -             &   BDLEELSSO          &  0.2M               \\
                { \textbf{Complex CNN}}\cite{ristea2021complex} &  -             &   4.51km            &  -             &  0.26             &   BDLEELSSO          &  0.2M               \\
                { \textbf{Fourier Transformer}}\cite{ge2024multi} &  -             &   3.77km            &  -             &  0.19             &   BDLEELSSO          &  0.2M               \\
                \hline
                \end{tabular}
                \caption{Location prediction performance}\label{Location-Pred-performance}
            \end{table}

            In addition to its exceptional performance in phase picking and location estimation, our FisH model also demonstrates strong capabilities in earthquake magnitude prediction. Similar to the location estimation task, we select the estimated magnitude at 20 seconds after the P-wave arrival time as the final magnitude estimation results.As shown in Table~\ref{Location-Pred-performance}, the FisH model achieves a mean absolute error (MAE) of 0.14 in magnitude estimation on the STEAD-BDLEELSSO dataset, which is the best among all the compared methods.

            Most compared consider phase picking, location estimation, magnitude estimation as independent task and perform each of them by using a separate network. By seamlessly integrating all these three tasks within its unified neural network architecture, the FisH model streamlines the overall earthquake monitoring process and enhances the overall performance of the early warning system.
    \subsection*{Real World Monitoring}
\begin{figure}[ht]
            \centering
            \includegraphics[width=\linewidth]{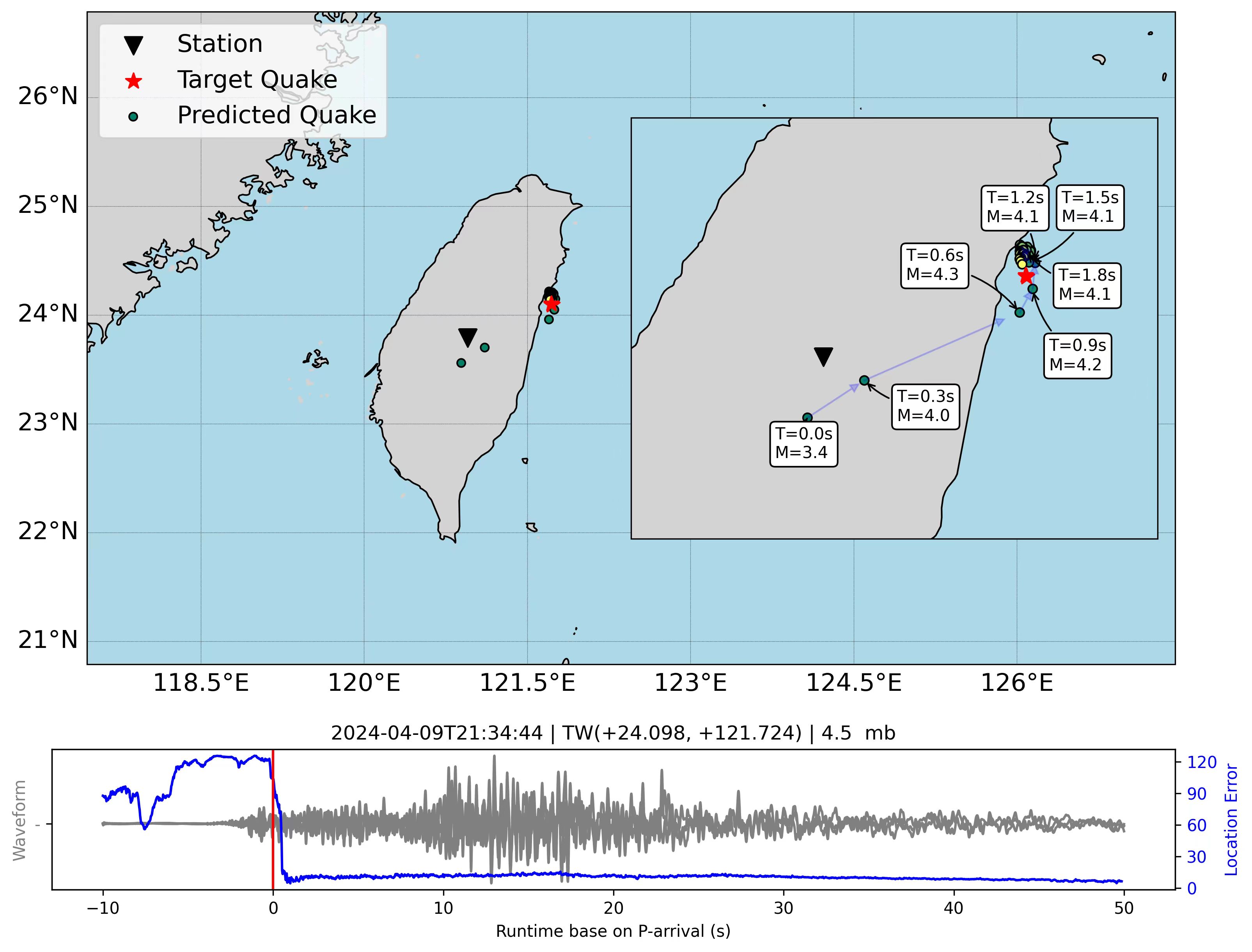}
            \caption{
            We apply the FisH model to the recent earthquake in Taiwan on April 9th \cite{irisSAGEWilber}. In this example, the warning system quickly locates the earthquake's position approximately 1 second after receiving the P-wave signal and provides an estimation of the event's magnitude. In the figure above, green nodes mark the predicted locations at each timestamp based on the P-wave arrival. The blue arrow represents the prediction trajectory. The lower part of the figure displays the location error from just after the P-wave arrival up to 50 seconds later, superimposed on a gray waveform curve as the background.
            }
            \label{fig:tw_simulation}
        \end{figure}

        To assess the real-world applicability of our model, we applied the FisH model to actual earthquake data collected from the SAGE website \cite{irisSAGEWilber}. In this scenario (as shown in Fig.\ref{fig:tw_simulation}), the warning system efficiently locates the earthquake's position approximately 1 second after detecting the P-wave signal and provides an estimation of the event's magnitude."

\section*{Discussion}
    
    \subsection*{The Rationale for Using RetNet over Vanilla Transformers}
        The choice to use RetNet instead of Transformers as Encoder module for inference is driven by several factors, with computational cost being a core consideration. RetNet's inference cost, for both computation and memory, is $O(1)$, making it highly efficient compared to Transformer's $O(n^2)$ inference cost, which becomes prohibitive for larger models. While smaller models with approximately 100M parameters or less, operating on shorter sequences (e.g., 6000), may not exhibit significant differences between RetNet and Transformer during GPU-based inference, larger models required for more extensive seismic data would be disproportionately impacted by Transformer's quadratic explosion of requirements. The RetNet architecture enables the deployment of real-time monitoring on small chips at the station, eliminating the need for post-earthquake analysis using supercomputers. Therefore, a solution combining a linear Transformer, controllable memory window, and zero-mode response, as embodied by RetNet, becomes the preferred choice.
        
    \subsection*{Key Features of Our Architecture: Enabling Monitoring, Warning, and Prediction as Mechanism Solutions}
        In the pursuit of developing a framework capable of dynamically adapting to 'post-processing', 'real-time', and 'prediction' tasks, two distinct strategies are considered: 'Large Model + Windowing' and 'Large Model + RNN'.

        The 'Large Model + Windowing' strategy, despite its high computational cost, is powerful for 'post-processing' and 'real-time warning' tasks given high-quality 'pre-shock', 'shock', and 'post-shock' data. However, this approach faces challenges in 'prediction' tasks, as the model must identify relevant 'shock' features from the 'noise' data variations to characterize earthquake occurrence and predict the timing and location of the earthquake. For this model to function effectively, there must be a discernible difference between 'pre-shock' data and 'noise' data.
        
        On the other hand, the 'Large Model + RNN' strategy can be seen as an 'any window' model. During training, different regions can be assigned distinct weights to focus the model's ability on specific areas. For example, the model can be trained to respond to the 'shock phase' in a window from the 'P-wave arrival' to the 'S-wave end', preventing the model from concerning itself with waveforms outside this area. When combined with designed structures such as "zero-response" and "decay memory", the 'Large Model + RNN' approach offers the ability to dynamically adapt to 'post-processing', 'real-time', and 'prediction' tasks within the same framework. As a result, challenging requirements such as 'immortal running', 'real-time warning', and 'continuous quake monitoring' become features under this mechanism rather than obstacles.
        \begin{itemize}
            \item When the model inference to the stamp before the P-arrival, it is in the `prediction' phase
            \item When the model inference to the near after the P-arrival, it is in the  `warning' phase
            \item When the model inference to the far away from the S-arrival, it is in the  `post-processing' phase
        \end{itemize}

    \subsection*{Unified Architecture}
        The FisH architecture, as demonstrated, offers a transformative approach by converting wave information into a unified embedding, enabling adaptation to various seismic sub-tasks such as picking, locating, and magnitude estimation. In contrast to previous approaches where separate workflows were designed for different seismic quantities, FisH streamlines the process by utilizing a single pretrained backbone and small downstream head for each specific task.
        
        Traditionally, researchers employed distinct methodologies, such as using normalized data for picking and P-surround data for locating, and building tailored expert models for each sub-task. However, within the FisH framework, once the backbone is pretrained, each task only requires a small downstream head. In this paper, we validate the performance of FisH on the picking, location, and magnitude tasks. However, the architecture can easily be extended to encompass more comprehensive quantities, such as 'Focal Mechanism,' as long as the necessary data is prepared.
        
        The real-time encoding of each timestamp by the FisH model facilitates straightforward association of multi-stations, enabling the creation of a large station-network shoal model. This capability opens up possibilities for leveraging multiple stations' data simultaneously, enhancing the model's performance and enabling a more comprehensive understanding of seismic events.
        
        In summary, the FisH architecture presents a unified and adaptable approach to seismic sub-tasks, eliminating the need for separate workflows and expert models. Its flexibility allows for easy extension to additional quantities beyond picking, location, and magnitude estimation. With its real-time encoding capabilities and multi-station association potential, FisH demonstrates the potential to revolutionize seismic analysis by providing a comprehensive and efficient solution for real-time monitoring and understanding of seismic events.

    \subsection*{Future Work}
        There are still some important future works on the agenda. For example, the STEAD and INSTANCE datasets used in this study are pre-normalized, which may not accurately represent real-time waveform flow. To fully validate the effectiveness of FisH, it is necessary to apply the method to real-end signal systems. Adapting the model to handle raw, non-normalized waveform data and testing its ability to perform real-time event detection and characterization is essential to ensure the model's practicality and reliability in real-world scenarios.
        
        Another important future direction is developing a multi-station FisH framework. By incorporating data from multiple seismic stations, the model can potentially achieve much better precision in event detection, location estimation, and characterization. This extension will involve addressing challenges related to data synchronization, fusion, and communication between stations, as well as adapting the model architecture to handle multi-station inputs effectively.
        
        Exploring the integration of complementary data sources, such as GPS, InSAR, or other geophysical measurements, can further improve the model's performance and provide a more comprehensive understanding of seismic events. This may help push the boundaries of early warning systems and quake prediction.
        
        Investigating the use of transfer learning and domain adaptation techniques can enhance the model's generalization ability across different geographical regions and seismic networks. This will enable the model's deployment in various seismic monitoring contexts and improve its practicality.

\section*{Conclusion}     
   In this study, we introduce FisH (Fast Information Streaming Handler), a novel machine learning model for real-time seismic event detection and characterization using single-station waveform data. Our key findings are as follows:

FisH demonstrates exceptional performance on the STEAD dataset in seismic phase picking, achieving 99\% precision and recall for P-arrival detection with minimal delay. The model also provides continuous and accurate earthquake magnitude estimates, reaching a minimum mean absolute error of 0.15 at 7 seconds after the P-wave arrival or 3 seconds after the S-arrival. Furthermore, FisH directly predicts the relative location of earthquakes, with the location error decreasing to around 6 km within 10 seconds after the P-arrival or 1 second after the S-arrival.

Importantly, FisH can handle continuous seismic data streams seamlessly, automatically resetting to its initial state after long periods of non-earthquake activity. Compared to previous work on individual tasks such as phase picking, location estimation, and magnitude estimation, FisH integrates these capabilities into a unified framework and achieves competitive or superior results across various benchmark datasets.

Our study also reveals several intriguing findings from a data science perspective when pushing the limits of machine learning algorithms. For instance, we uncover the "information boundary" that limits the earliest possible warning time for earthquake early warning systems, the power-law distribution of location prediction errors, and the extremely challenging nature of P-wave-based early predictions given the current dataset.

In conclusion, FisH demonstrates the potential of machine learning for real-time seismic monitoring using single-station waveform data. Its ability to perform multiple tasks simultaneously and handle continuous data streams makes it a promising tool for practical earthquake early warning applications.


\bibliography{sample}

\begin{thebibliography}{10}
\urlstyle{rm}
\expandafter\ifx\csname url\endcsname\relax
  \def\url#1{\texttt{#1}}\fi
\expandafter\ifx\csname urlprefix\endcsname\relax\def\urlprefix{URL }\fi
\expandafter\ifx\csname doiprefix\endcsname\relax\def\doiprefix{DOI: }\fi
\providecommand{\bibinfo}[2]{#2}
\providecommand{\eprint}[2][]{\url{#2}}

\bibitem{allen2009real}
\bibinfo{author}{Allen, R.~M.} \emph{et~al.}
\newblock \bibinfo{journal}{\bibinfo{title}{Real-time earthquake detection and hazard assessment by elarms across california}}.
\newblock {\emph{\JournalTitle{Geophysical Research Letters}}} \textbf{\bibinfo{volume}{36}} (\bibinfo{year}{2009}).

\bibitem{wurman2007toward}
\bibinfo{author}{Wurman, G.}, \bibinfo{author}{Allen, R.~M.} \& \bibinfo{author}{Lombard, P.}
\newblock \bibinfo{journal}{\bibinfo{title}{Toward earthquake early warning in northern california}}.
\newblock {\emph{\JournalTitle{Journal of Geophysical Research: Solid Earth}}} \textbf{\bibinfo{volume}{112}} (\bibinfo{year}{2007}).

\bibitem{serdar2014designing}
\bibinfo{author}{Serdar~Kuyuk, H.} \emph{et~al.}
\newblock \bibinfo{journal}{\bibinfo{title}{Designing a network-based earthquake early warning algorithm for california: Elarms-2}}.
\newblock {\emph{\JournalTitle{Bulletin of the Seismological Society of America}}} \textbf{\bibinfo{volume}{104}}, \bibinfo{pages}{162--173} (\bibinfo{year}{2014}).

\bibitem{chung2019optimizing}
\bibinfo{author}{Chung, A.~I.}, \bibinfo{author}{Henson, I.} \& \bibinfo{author}{Allen, R.~M.}
\newblock \bibinfo{journal}{\bibinfo{title}{Optimizing earthquake early warning performance: Elarms-3}}.
\newblock {\emph{\JournalTitle{Seismological Research Letters}}} \textbf{\bibinfo{volume}{90}}, \bibinfo{pages}{727--743} (\bibinfo{year}{2019}).

\bibitem{chen2015earthworm}
\bibinfo{author}{Chen, D.-Y.}, \bibinfo{author}{Hsiao, N.-C.} \& \bibinfo{author}{Wu, Y.-M.}
\newblock \bibinfo{journal}{\bibinfo{title}{The earthworm based earthquake alarm reporting system in taiwan}}.
\newblock {\emph{\JournalTitle{Bulletin of the Seismological Society of America}}} \textbf{\bibinfo{volume}{105}}, \bibinfo{pages}{568--579} (\bibinfo{year}{2015}).

\bibitem{hsiao2009development}
\bibinfo{author}{Hsiao, N.-C.}, \bibinfo{author}{Wu, Y.-M.}, \bibinfo{author}{Shin, T.-C.}, \bibinfo{author}{Zhao, L.} \& \bibinfo{author}{Teng, T.-L.}
\newblock \bibinfo{journal}{\bibinfo{title}{Development of earthquake early warning system in taiwan}}.
\newblock {\emph{\JournalTitle{Geophysical research letters}}} \textbf{\bibinfo{volume}{36}} (\bibinfo{year}{2009}).

\bibitem{wu2002virtual}
\bibinfo{author}{Wu, Y.-M.} \& \bibinfo{author}{Teng, T.-l.}
\newblock \bibinfo{journal}{\bibinfo{title}{A virtual subnetwork approach to earthquake early warning}}.
\newblock {\emph{\JournalTitle{Bulletin of the Seismological Society of America}}} \textbf{\bibinfo{volume}{92}}, \bibinfo{pages}{2008--2018} (\bibinfo{year}{2002}).

\bibitem{peng2011developing}
\bibinfo{author}{Peng, H.} \emph{et~al.}
\newblock \bibinfo{journal}{\bibinfo{title}{Developing a prototype earthquake early warning system in the beijing capital region}}.
\newblock {\emph{\JournalTitle{Seismological Research Letters}}} \textbf{\bibinfo{volume}{82}}, \bibinfo{pages}{394--403} (\bibinfo{year}{2011}).

\bibitem{allen2003potential}
\bibinfo{author}{Allen, R.~M.} \& \bibinfo{author}{Kanamori, H.}
\newblock \bibinfo{journal}{\bibinfo{title}{The potential for earthquake early warning in southern california}}.
\newblock {\emph{\JournalTitle{Science}}} \textbf{\bibinfo{volume}{300}}, \bibinfo{pages}{786--789} (\bibinfo{year}{2003}).

\bibitem{tsang2007magnitude}
\bibinfo{author}{Tsang, L.~L.}, \bibinfo{author}{Allen, R.~M.} \& \bibinfo{author}{Wurman, G.}
\newblock \bibinfo{journal}{\bibinfo{title}{Magnitude scaling relations from p-waves in southern california}}.
\newblock {\emph{\JournalTitle{Geophysical Research Letters}}} \textbf{\bibinfo{volume}{34}} (\bibinfo{year}{2007}).

\bibitem{wu2008development}
\bibinfo{author}{Wu, Y.-M.} \& \bibinfo{author}{Kanamori, H.}
\newblock \bibinfo{journal}{\bibinfo{title}{Development of an earthquake early warning system using real-time strong motion signals}}.
\newblock {\emph{\JournalTitle{Sensors}}} \textbf{\bibinfo{volume}{8}}, \bibinfo{pages}{1--9} (\bibinfo{year}{2008}).

\bibitem{wu2008exploring}
\bibinfo{author}{Wu, Y.-M.} \& \bibinfo{author}{Kanamori, H.}
\newblock \bibinfo{journal}{\bibinfo{title}{Exploring the feasibility of on-site earthquake early warning using close-in records of the 2007 noto hanto earthquake}}.
\newblock {\emph{\JournalTitle{Earth, Planets and Space}}} \textbf{\bibinfo{volume}{60}}, \bibinfo{pages}{155--160} (\bibinfo{year}{2008}).

\bibitem{wu2006magnitude}
\bibinfo{author}{Wu, Y.-M.} \& \bibinfo{author}{Zhao, L.}
\newblock \bibinfo{journal}{\bibinfo{title}{Magnitude estimation using the first three seconds p-wave amplitude in earthquake early warning}}.
\newblock {\emph{\JournalTitle{Geophysical research letters}}} \textbf{\bibinfo{volume}{33}} (\bibinfo{year}{2006}).

\bibitem{cua2007virtual}
\bibinfo{author}{Cua, G.} \& \bibinfo{author}{Heaton, T.}
\newblock \bibinfo{title}{The virtual seismologist (vs) method: A bayesian approach to earthquake early warning}.
\newblock In \emph{\bibinfo{booktitle}{Earthquake early warning systems}}, \bibinfo{pages}{97--132} (\bibinfo{publisher}{Springer}, \bibinfo{year}{2007}).

\bibitem{cua2009real}
\bibinfo{author}{Cua, G.}, \bibinfo{author}{Fischer, M.}, \bibinfo{author}{Heaton, T.} \& \bibinfo{author}{Wiemer, S.}
\newblock \bibinfo{journal}{\bibinfo{title}{Real-time performance of the virtual seismologist earthquake early warning algorithm in southern california}}.
\newblock {\emph{\JournalTitle{Seismological Research Letters}}} \textbf{\bibinfo{volume}{80}}, \bibinfo{pages}{740--747} (\bibinfo{year}{2009}).

\bibitem{bose2007earthquake}
\bibinfo{author}{B{\"o}se, M.}, \bibinfo{author}{Ionescu, C.} \& \bibinfo{author}{Wenzel, F.}
\newblock \bibinfo{journal}{\bibinfo{title}{Earthquake early warning for bucharest, romania: Novel and revised scaling relations}}.
\newblock {\emph{\JournalTitle{Geophysical Research Letters}}} \textbf{\bibinfo{volume}{34}} (\bibinfo{year}{2007}).

\bibitem{muarmureanu2011advanced}
\bibinfo{author}{M{\u{a}}rmureanu, A.}, \bibinfo{author}{Ionescu, C.} \& \bibinfo{author}{Cioflan, C.}
\newblock \bibinfo{journal}{\bibinfo{title}{Advanced real-time acquisition of the vrancea earthquake early warning system}}.
\newblock {\emph{\JournalTitle{Soil Dynamics and Earthquake Engineering}}} \textbf{\bibinfo{volume}{31}}, \bibinfo{pages}{163--169} (\bibinfo{year}{2011}).

\bibitem{ionescu2007early}
\bibinfo{author}{Ionescu, C.} \emph{et~al.}
\newblock \bibinfo{journal}{\bibinfo{title}{An early warning system for deep vrancea (romania) earthquakes}}.
\newblock {\emph{\JournalTitle{Earthquake early warning systems}}} \bibinfo{pages}{343--349} (\bibinfo{year}{2007}).

\bibitem{sheen2014magnitude}
\bibinfo{author}{Sheen, D.-H.}, \bibinfo{author}{Lim, I.-S.}, \bibinfo{author}{Park, J.-H.} \& \bibinfo{author}{Chi, H.-C.}
\newblock \bibinfo{journal}{\bibinfo{title}{Magnitude scaling relationships using p waves for earthquake early warning in south korea}}.
\newblock {\emph{\JournalTitle{Geosciences Journal}}} \textbf{\bibinfo{volume}{18}}, \bibinfo{pages}{7--12} (\bibinfo{year}{2014}).

\bibitem{sheen2017first}
\bibinfo{author}{Sheen, D.-H.} \emph{et~al.}
\newblock \bibinfo{journal}{\bibinfo{title}{The first stage of an earthquake early warning system in south korea}}.
\newblock {\emph{\JournalTitle{Seismological Research Letters}}} \textbf{\bibinfo{volume}{88}}, \bibinfo{pages}{1491--1498} (\bibinfo{year}{2017}).

\bibitem{carranza2013earthquake}
\bibinfo{author}{Carranza, M.}, \bibinfo{author}{Buforn, E.}, \bibinfo{author}{Colombelli, S.} \& \bibinfo{author}{Zollo, A.}
\newblock \bibinfo{journal}{\bibinfo{title}{Earthquake early warning for southern iberia: Ap wave threshold-based approach}}.
\newblock {\emph{\JournalTitle{Geophysical research letters}}} \textbf{\bibinfo{volume}{40}}, \bibinfo{pages}{4588--4593} (\bibinfo{year}{2013}).

\bibitem{zhu2019phasenet}
\bibinfo{author}{Zhu, W.} \& \bibinfo{author}{Beroza, G.~C.}
\newblock \bibinfo{journal}{\bibinfo{title}{Phasenet: a deep-neural-network-based seismic arrival-time picking method}}.
\newblock {\emph{\JournalTitle{Geophysical Journal International}}} \textbf{\bibinfo{volume}{216}}, \bibinfo{pages}{261--273} (\bibinfo{year}{2019}).

\bibitem{mousavi2020earthquake}
\bibinfo{author}{Mousavi, S.~M.}, \bibinfo{author}{Ellsworth, W.~L.}, \bibinfo{author}{Zhu, W.}, \bibinfo{author}{Chuang, L.~Y.} \& \bibinfo{author}{Beroza, G.~C.}
\newblock \bibinfo{journal}{\bibinfo{title}{Earthquake transformer—an attentive deep-learning model for simultaneous earthquake detection and phase picking}}.
\newblock {\emph{\JournalTitle{Nature communications}}} \textbf{\bibinfo{volume}{11}}, \bibinfo{pages}{3952} (\bibinfo{year}{2020}).

\bibitem{mousavi2020bayesian}
\bibinfo{author}{Mousavi, S.~M.} \& \bibinfo{author}{Beroza, G.~C.}
\newblock \bibinfo{journal}{\bibinfo{title}{Bayesian-deep-learning estimation of earthquake location from single-station observations}}.
\newblock {\emph{\JournalTitle{IEEE Transactions on Geoscience and Remote Sensing}}} \textbf{\bibinfo{volume}{58}}, \bibinfo{pages}{8211--8224} (\bibinfo{year}{2020}).

\bibitem{mousavi2020machine}
\bibinfo{author}{Mousavi, S.~M.} \& \bibinfo{author}{Beroza, G.~C.}
\newblock \bibinfo{journal}{\bibinfo{title}{A machine-learning approach for earthquake magnitude estimation}}.
\newblock {\emph{\JournalTitle{Geophysical Research Letters}}} \textbf{\bibinfo{volume}{47}}, \bibinfo{pages}{e2019GL085976} (\bibinfo{year}{2020}).

\bibitem{sun2023retentive}
\bibinfo{author}{Sun, Y.} \emph{et~al.}
\newblock \bibinfo{journal}{\bibinfo{title}{Retentive network: A successor to transformer for large language models}}.
\newblock {\emph{\JournalTitle{arXiv preprint arXiv:2307.08621}}}  (\bibinfo{year}{2023}).

\bibitem{mousavi2019stanford}
\bibinfo{author}{Mousavi, S.~M.}, \bibinfo{author}{Sheng, Y.}, \bibinfo{author}{Zhu, W.} \& \bibinfo{author}{Beroza, G.~C.}
\newblock \bibinfo{journal}{\bibinfo{title}{Stanford earthquake dataset (stead): A global data set of seismic signals for ai}}.
\newblock {\emph{\JournalTitle{IEEE Access}}}  (\bibinfo{year}{2019}).

\bibitem{su2024roformer}
\bibinfo{author}{Su, J.} \emph{et~al.}
\newblock \bibinfo{journal}{\bibinfo{title}{Roformer: Enhanced transformer with rotary position embedding}}.
\newblock {\emph{\JournalTitle{Neurocomputing}}} \textbf{\bibinfo{volume}{568}}, \bibinfo{pages}{127063} (\bibinfo{year}{2024}).

\bibitem{accelerate}
\bibinfo{author}{Gugger, S.} \emph{et~al.}
\newblock \bibinfo{title}{Accelerate: Training and inference at scale made simple, efficient and adaptable.}
\newblock \bibinfo{howpublished}{\url{https://github.com/huggingface/accelerate}} (\bibinfo{year}{2022}).

\bibitem{rasley2020deepspeed}
\bibinfo{author}{Rasley, J.}, \bibinfo{author}{Rajbhandari, S.}, \bibinfo{author}{Ruwase, O.} \& \bibinfo{author}{He, Y.}
\newblock \bibinfo{title}{Deepspeed: System optimizations enable training deep learning models with over 100 billion parameters}.
\newblock In \emph{\bibinfo{booktitle}{Proceedings of the 26th ACM SIGKDD International Conference on Knowledge Discovery \& Data Mining}}, \bibinfo{pages}{3505--3506} (\bibinfo{year}{2020}).

\bibitem{michelini2021instance}
\bibinfo{author}{Michelini, A.} \emph{et~al.}
\newblock \bibinfo{journal}{\bibinfo{title}{Instance--the italian seismic dataset for machine learning}}.
\newblock {\emph{\JournalTitle{Earth System Science Data}}} \textbf{\bibinfo{volume}{13}}, \bibinfo{pages}{5509--5544} (\bibinfo{year}{2021}).

\bibitem{10.1785/0220210309}
\bibinfo{author}{Yu, Z.} \& \bibinfo{author}{Wang, W.}
\newblock \bibinfo{journal}{\bibinfo{title}{{LPPN: A Lightweight Network for Fast Phase Picking}}}.
\newblock {\emph{\JournalTitle{Seismological Research Letters}}} \textbf{\bibinfo{volume}{93}}, \bibinfo{pages}{2834--2846}, \doiprefix\url{10.1785/0220210309} (\bibinfo{year}{2022}).
\newblock \eprint{https://pubs.geoscienceworld.org/ssa/srl/article-pdf/93/5/2834/5682120/srl-2021309.1.pdf}.

\bibitem{li2022epick}
\bibinfo{author}{Li, W.} \emph{et~al.}
\newblock \bibinfo{journal}{\bibinfo{title}{Epick: Attention-based multi-scale unet for earthquake detection and seismic phase picking}}.
\newblock {\emph{\JournalTitle{Frontiers in Earth Science}}} \textbf{\bibinfo{volume}{10}}, \bibinfo{pages}{953007} (\bibinfo{year}{2022}).

\bibitem{9467532}
\bibinfo{author}{Saad, O.~M.} \& \bibinfo{author}{Chen, Y.}
\newblock \bibinfo{journal}{\bibinfo{title}{Capsphase: Capsule neural network for seismic phase classification and picking}}.
\newblock {\emph{\JournalTitle{IEEE Transactions on Geoscience and Remote Sensing}}} \textbf{\bibinfo{volume}{60}}, \bibinfo{pages}{1--11}, \doiprefix\url{10.1109/TGRS.2021.3089929} (\bibinfo{year}{2022}).

\bibitem{meng2024seisparanet}
\bibinfo{author}{Meng, F.}, \bibinfo{author}{Ren, T.}, \bibinfo{author}{Zhang, H.}, \bibinfo{author}{Wang, X.} \& \bibinfo{author}{Chen, H.}
\newblock \bibinfo{journal}{\bibinfo{title}{Seisparanet: A novel multi-task network for seismic source characterization in earthquake early warning}}.
\newblock {\emph{\JournalTitle{IEEE Transactions on Geoscience and Remote Sensing}}}  (\bibinfo{year}{2024}).

\bibitem{ristea2021complex}
\bibinfo{author}{Ristea, N.-C.} \& \bibinfo{author}{Radoi, A.}
\newblock \bibinfo{journal}{\bibinfo{title}{Complex neural networks for estimating epicentral distance, depth, and magnitude of seismic waves}}.
\newblock {\emph{\JournalTitle{IEEE Geoscience and Remote Sensing Letters}}} \textbf{\bibinfo{volume}{19}}, \bibinfo{pages}{1--5} (\bibinfo{year}{2021}).

\bibitem{ge2024multi}
\bibinfo{author}{Ge, K.}, \bibinfo{author}{Wang, C.}, \bibinfo{author}{Guo, Y.-T.}, \bibinfo{author}{Tang, Y.-S.} \& \bibinfo{author}{Fan, J.-S.}
\newblock \bibinfo{journal}{\bibinfo{title}{A multi-task fourier transformer network for seismic source characterization estimation from a single-station waveform}}.
\newblock {\emph{\JournalTitle{IEEE Geoscience and Remote Sensing Letters}}}  (\bibinfo{year}{2024}).

\bibitem{irisSAGEWilber}
\bibinfo{title}{{S}{A}{G}{E}: {W}ilber 3: {S}elect {S}tations --- ds.iris.edu}.
\newblock \bibinfo{howpublished}{\url{https://ds.iris.edu/wilber3/find_stations/11826724}}.
\newblock \bibinfo{note}{[Accessed 23-04-2024]}.

\end{thebibliography}
\end{document}